\pdfoutput=1

\documentclass[11pt]{article}

\usepackage[preprint]{acl}

\usepackage{times}
\usepackage{latexsym}

\usepackage[T1]{fontenc}

\usepackage[utf8]{inputenc}

\usepackage{microtype}

\usepackage{inconsolata}

\usepackage{graphicx}

\usepackage{booktabs, xspace,makecell, multirow}
\usepackage{enumitem, amssymb}
\usepackage{float}

\usepackage{listings} %

\newcommand{\fen}{F\textsubscript{1}\xspace}

\newcommand{\xlmr}{XLM-R\xspace} %
\newcommand{\wrt}{with respect to\xspace}
\newcommand{\cgpt}{ChatGPT\xspace}
\setlist{nosep, leftmargin=*, labelsep=0.5em, itemsep=0.25em, before=\vspace{0.2em}, after=\vspace{0.2em}}

\title{LTG at SemEval-2025 Task 10: Optimizing Context for Classification of Narrative Roles}

\author{Egil Rønningstad \\
  Department of Informatics  \\
  University of Oslo, Norway \\
  \texttt{egilron@uio.no}  \\\And
  Gaurav Negi  \\
  Data Science Institute  \\
 University of Galway \\
   \texttt{gaurav.negi@insight-centre.org} \\
  }

\begin{document}
\maketitle
\begin{abstract}
Our contribution to the SemEval 2025 shared task 10, subtask 1 on entity framing, tackles the challenge of providing the necessary segments from longer documents as context for classification with a masked language model. We show that a simple entity-oriented heuristics for context selection can enable text classification using models with limited context window. Our context selection approach 
and the XLM-RoBERTa language model is on par with, or outperforms, Supervised Fine-Tuning with larger generative language models.
\end{abstract}

\section{Introduction}

Shared task 10, subtask 1 for SemEval 2025 \citep{semeval2025task10} presents annotated news articles in Bulgarian, English, Hindi, (European) Portuguese, and Russian. For each article, a number of entities are identified and annotated for narrative roles. These entities are in the text framed as playing one of three main narrative roles; protagonists, antagonists or innocent. For each main role, there is a number of fine-grained roles, 22 in total. Each entity may be given multiple fine-grained roles available for the assigned main role. Further details on the task can be found in the technical report \citep{guidelinesSE25T10}. By classifying how entities are framed in news articles, we can better understand reporting angles and identify bias variations between different news sources.

Our contribution focuses on the following challenges in modeling the narrative roles of the annotated entities: 
\begin{enumerate} [label=\alph*)]
    \item Each article may contain much irrelevant text \wrt a given entity.
    \item A sentence mentioning an entity may also mention other entities and frame each differently.
    \item The text segment(s) contributing to the entity framing may span multiple sentences.
\end{enumerate}
As seen in Table \ref{tab:train_test_size}, the provided training examples are in the hundreds and the thousands for each language. We therefore hypothesize that XLM-RoBERTa-large (\xlmr), a multilingual masked language model, could be a cost-effective starting point. Due to this, we pose the following research questions:
\begin{itemize} 
    \item \textbf{RQ1:} Can we find rule-based approaches that mitigate the above mentioned challenges, including irrelevant text and documents that will not fit in the \xlmr context window?
    \item \textbf{RQ2:} Can a fine-tuned \xlmr-based model be outperformed by larger language models where the entire document fits well within the context window?
\end{itemize}

To answer these questions, our paper presents results from experiments regarding text pre-processing where we fine-tune \xlmr on various text segments and evaluate each pre-processing strategy. These results are compared against zero-shot prompting of a large language model (LLM), and Supervised Fine-Tuning of LLMs with 7-8 billion parameters.

\subsection{Context Optimization}
Extracting only the relevant text segments for a given task can be named ``context optimization''. Studies have shown that for long-context LLMs, irrelevant or distractive text as part of an input, reduces model performance \citep{shi2023large, wu2024reducing, cai-etal-2024-joint, wang2024fact}. This task is also being described as ``context rewriting'' \citep{wang2024fact} or ``prompt compression'' \citep{liskavets2024prompt}, but for this paper we prefer the term ``context optimization''. For MLMs this task is imperative due to the limited context window of, as for \xlmr, 512 subword tokens, which is not enough for many texts longer than microblog messages and short user-submitted reviews. 

\subsection{Supervised Fine-Tuning}
Supervised Fine-Tuning (SFT) has proven to to be a viable path for creating text classification models. 
Depending on the training examples and compute resources available, this approach may lead to better classification than in-context learning where pairs of example text and labels are fed to a LLM before requesting the classification of a new example \cite{mosbach-etal-2023-shot}. 

\subsection{Dataset}
The data for this shared task were released sequentially. The experiments reported here, are trained with the finalized train split and evaluated on the labeled dev split. The number of labeled entities for training and evaluation is found in Table \ref{tab:train_test_size}.  

\begin{table}[ht] %
    \centering
\begin{tabular}{lrr}
\toprule
 
Language & train  & dev  \\
\midrule
BG & 625 & 30 \\
EN & 686 & 91 \\
HI & 2331 & 280 \\
PT & 1245 & 116 \\
RU & 366 & 86 \\
\midrule
all & 5253 & 603 \\
\bottomrule
\end{tabular}
    \caption{Annotated entities per language in the train and dev splits for the experiments reported on in this paper. }
    \label{tab:train_test_size}
\end{table}
\begin{figure*}[t]
    \centering
    \includegraphics[width=0.8\textwidth, height=0.53\textwidth]{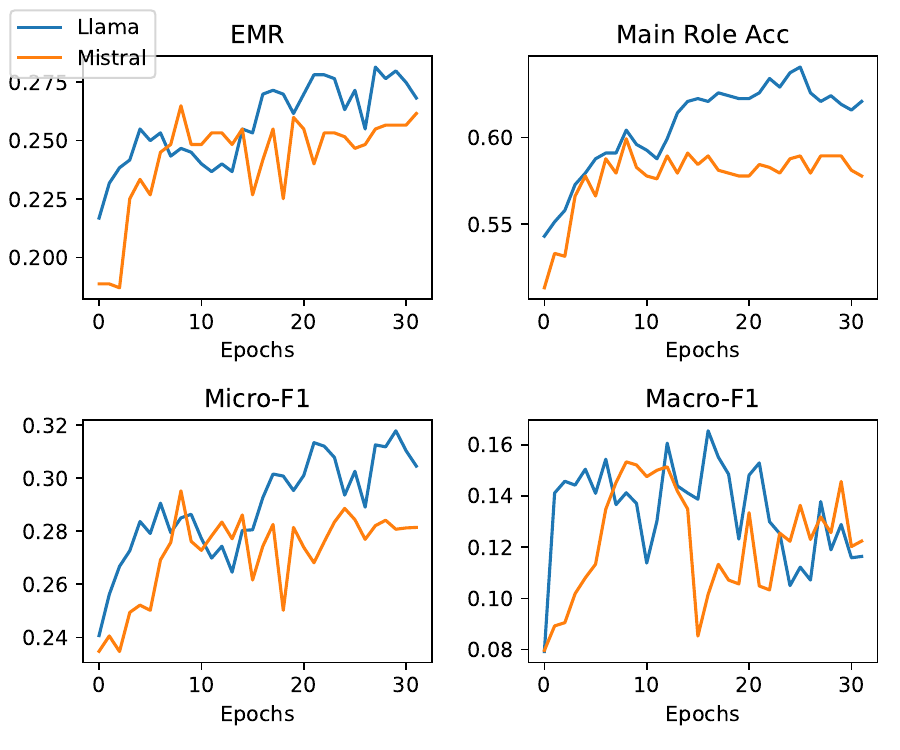}
    \caption{SFT learning trends per epoch as evaluated on dev dataset}
    \label{fig:eval_metrics_training}
\end{figure*}

\section{Dataset Pre-Processing} \label{sec:dataset-preproc}
Our contribution to the shared task was trained exclusively on the provided data. 
To address the above mentioned challenges for narrative role classification and answer RQ1, we experimented with how to best prepare the data for fine-tuning and classification.

\subsection{Text Span Extraction} \label{sec:span_extract}
We hypothesize that the text relevant to the entity framing would be located in proximity to the entity mention. As the provided texts are split in sentences and paragraph, and the entities in question are pre-identified within the text, we compare a selection of rule-based context extraction approaches, and measure their performance against a simple LLM-generated baseline. To assess the value of such text span extraction, we performed experiments with the following alternative text extraction heuristics:
\begin{enumerate} [label=\alph*)]
    \item \textbf{Single sentences} For each annotated entity, provide only the sentence where the entity is mentioned.
    \item \textbf{Single paragraph} For each annotated entity, provide the paragraph in which the entity occurs.
    \item \textbf{Entire text} For each annotated entity, provide the entire document. The model consumes as much text as possible, ignoring the rest.
    \item \textbf{Entity-to-entity (ent2ent)} For each annotated entity, provide the sentence where the entity is mentioned, and all subsequent sentences until a new entity occurs.
    \item \textbf{GPT-extracted} For each annotated entity, provide the replies through the api of \cgpt with gpt-4o, queried to extract the text span(s) containing information regarding the narrative role of the entity.
\end{enumerate}

\subsection{Merging languages} 
When it comes to classification tasks with a multilingual pretrained model, there is the tradeoff between getting a larger training set, and introducing ``noise'' from the language variety \citep{conneau-etal-2020-unsupervised, ronningstad-2023-uio}. We therefore prepared one dataset per language, and one merged dataset containing all languages.

\subsection{Preparing data for SFT}
We prepared one prompt per text document for Supervised Fine-Tuning.

\paragraph{Prompt Template}
The prompt template (see Appendix \ref{appendix:nlprompts}) used for making the predictions consists of the following segments: (i) Annotation Instructions, (ii) Taxonomy of the primary and secondary entity roles, (iii) The definitions of primary roles, (iv) Document input along with the entities, and (v) Output format.

It should be noted that the entities in question, occur multiple times in a few documents. To localize the correct instance of these entities an entity tag (<entity> </entity>) is placed around the occurrence using the index values provided in the annotations. This processed document is then included in the prompt.

\section{Modeling} \label{sec:modeling}

\begin{table*}[ht] %
    \centering
\begin{tabular}{llrrrrrr}
\toprule
Model & Method  & BG & EN & HI & PT & RU & All \\
\midrule
\multirowcell{6}{\xlmr} & ent2ent &  25.40 & 31.25 & 46.49 & 70.00 & \textbf{47.73} & \textbf{47.75} \\
 & ent2ent\_noprefix &  15.87 & 11.46 & 27.07 & 58.92 & 26.29 & 30.11 \\
 & ent2ent\_main2fine &  25.40 & 25.00 & 41.75 & \textbf{74.69} & 40.68 & 44.51 \\

 & sentence &  31.75 & 20.73 & \textbf{46.88} & 70.83 & 42.46 & 46.06 \\
 & gpt-extracted &  25.40 & 17.71 & 45.36 & 64.73 & 40.00 & 43.14 \\
 & paragraph &  25.40 & 16.75 & 40.00 & 67.50 & 38.64 & 40.79 \\
 & fulltext &  28.57 & 9.42 & 40.53 & 62.50 & 37.29 & 38.96 \\
 \midrule
 \cgpt-4o& gpt-inference &  38.96 & \textbf{36.95} & 36.85 & 65.64 & 34.65 & 41.78 \\
 \midrule

llama & SFT &\textbf{43.48} & 22.33 & 20.57 & 50.00 & 29.70 & 31.78\\
mistral & SFT & 36.36 & 30.77 & 18.71 & 49.17 & 34.83 & 29.52\\
\bottomrule
\end{tabular}
    \caption{Evaluation results for the various text extraction and modeling strategies for each test language. Evaluation metric is Micro \fen for the 22 fine-grained roles. We see that fine-tuning \xlmr on prefixed text segments using the \textit{ent2ent} segment extraction strategy yields the best overall results. All \xlmr experiments contain prefixed segments except the \textit{ent2ent\_noprefix} ablation experiment. All experiments classify the 22 fine-grained roles directly, except the \textit{ent2ent\_main2fine} experiment.
    }
    \label{tab:xlmr_contexts}
\end{table*}

We here present the various modeling approaches tested, and their evaluation results. The focus is on answering the research questions by applying the various text pre-processing approaches as input for \xlmr fine-tuning. These approaches are compared with classification based on larger language models.

\subsection{Modeling with \xlmr}
\xlmr-large\footnote{FacebookAI/xlm-roberta-large} was used for all \xlmr experiments.
No hyperparameters were altered from the standard settings for text classification in the Transformers library. Models were trained for 10 epochs.

\paragraph{Prepend Entity Mention.} 
The text spans provided to the model may be identical when classifying \wrt two different entities, and we therefore prepended the extracted segment with the prefix \textit{"Regarding <entity> :\textbackslash n"}, where <entity> is placeholder for the entity mention as written in the task annotations. This was done for all \xlmr experiments except for \textit{ent2ent\_noprefix}.

We see in Table \ref{tab:xlmr_contexts}, how the micro \fen scores were dramatically reduced when there was no prefix presenting the entity in question.

\paragraph{Monolingual vs Multilingual Fine-Tuning.}  
As can be seen in Table \ref{tab:train_language_vs_all} and Figure \ref{fig:trainset-contextresults} in Appendix \ref{appendix:languagewise},  fine-tuning on all languages improved results noticeably. For the languages with the smallest training set, there were no measurable Micro \fen results.  For subsequent experiments, the training data is understood to consist of all languages in the shared task.

\paragraph{Modeling Main Roles First}
As there are 22 fine-grained roles in total, and an entity may be labeled with multiple fine-grained roles, we attempted to employ the best-so-far method (Entity-to-entity with prefix) in a two-step modeling and inference approach (main2fine). We first trained a classifier for the three main roles, and predicted main roles for the dev set (one per entity). After separating the dev set into each predicted main role, we trained one multilabel classifier per predicted main role. This reduced the number of possible fine-grained roles to 6 for the Protagonist main role, 12 for Antagonist and 4 for 
Innocent. The results are found in Table \ref{tab:xlmr_contexts}, the row \textit{ent2ent\_main2fine}.

\subsection{Prompting an LLM for Classification} \label{sec:gpt-inference}
The results from the \xlmr -based models were compared against the labels provided by \cgpt-4o when queried for classifying one entity at a time, including the entire article and the label definitions in the prompt, but not any examples. The results are found under \textit{gpt-inference} in Table \ref{tab:xlmr_contexts}.

\subsection{Modeling with LLMs and LoRA}

 For these experiments, we used the instruction-tuned versions of Llama-3.1-8B\footnote{meta-llama/Llama-3.1-8B-Instruct} and Mistral-7B\footnote{mistralai/Mistral-7B-Instruct-v0.3} models. These models were further fine-tuned with the training dataset that was provided.

We use Low-Rank Adaptation (LoRA) for the parameter-efficient fine-tuning using the causal language modeling objective. The values prescribed for LoRA parameters in the introductory work were used ($r=32$ and $\alpha=64$). The parameters of the models (Llama and Mistral) were loaded with a 4-bit precision configuration. Since the models have multi-lingual capabilities, we combined the training dataset across all languages for fine-tuning. 

We fine-tuned LLMs for 32 epochs and selected the checkpoint with the highest Exact Match Ratio (EMR) to make the predictions. We track the metrics on the language-combined development dataset in Figure \ref{fig:eval_metrics_training}. Per-language final evaluation results are found in Table \ref{tab:xlmr_contexts}.

\begin{table}
    \centering
\begin{tabular}{lrrr}
\toprule
language & all & in-lang & samples \\
\midrule
RU & 52.33 & 0.00 & 366 \\
BG & 26.67 & 0.00 & 625 \\
EN & 29.67 & 8.79 & 686 \\
PT & 69.83 & 44.83 & 1245 \\
HI & 43.21 & 22.50 & 2331 \\
\midrule
all & 46.77 &  & 5253 \\
\bottomrule
\end{tabular}
    \caption{A comparison of test results (Micro \fen) with \xlmr and the ent2ent context optimization, when training either only on the training data of the test split language (in-lang) with their samples count (samples), or training on the entire train set (all). More results are presented in Figure \ref{fig:trainset-contextresults} in Appendix \ref{appendix:languagewise}.}
    \label{tab:train_language_vs_all}
\end{table}

\section{Analyses}
We here present our reflections on the results reported in the paper.

\subsection{Finding the Best Text Segments}
Table \ref{tab:xlmr_contexts} shows that modeling with \xlmr and the ent2ent-approach for segment extraction yields the best overall results. Although this approach is the best alternative for only one of the languages (Russian), we submit these results for all languages. Among our experiments, the ent2ent-approach yields competitive results for all languages except for Bulgarian where the experiments based on larger models are noticeably better.
 Our contribution ranked from fifth (Hindi and Portuguese) to tenth (English) on the official SemEval results on the test set.

\subsection{Performance vs train split size} 
\begin{figure}
    \centering
    \includegraphics[width=0.93\linewidth]{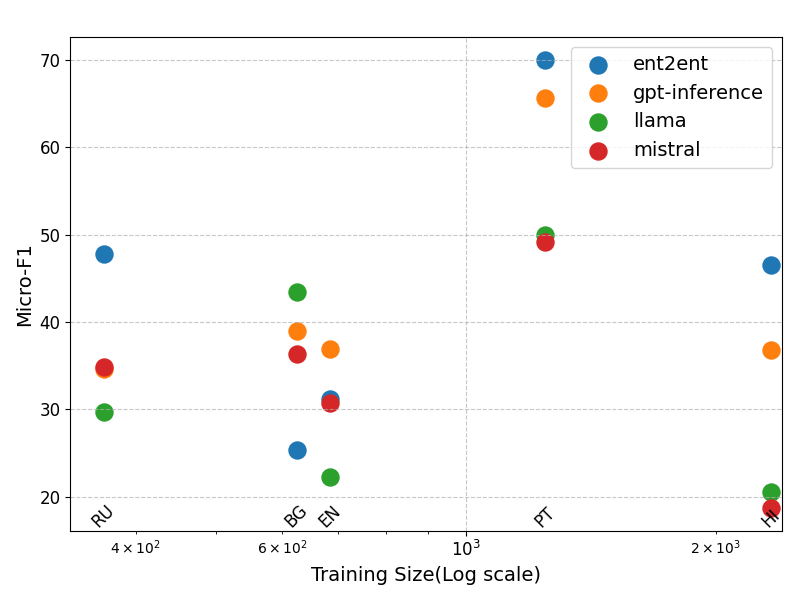}
    \caption{Model performance as a function of in-language training data}
    \label{fig:tr_size_microf1}
\end{figure}

All our approaches yielded large variations in results between languages. The training data as presented in Table \ref{tab:train_test_size} are very different in size per language. Although models were trained on the entire dataset, having data from the same language as the test data (in-language training data) has from experience proven to be important. We therefor compare the results with the size of the in-language train split. Figure \ref{fig:tr_size_microf1} shows no clear trend, as Portuguese yielded the best results, while Hindi had almost twice the amount of training data. We see that the \xlmr-based approach is clearly best for the two languages with the most training data. The need for more than a thousand in-language, in-domain training samples when fine-tuning \xlmr, is in line with previous experience. But again, \xlmr performs best of our approaches for Russian as well, with the fewest training samples. We can assume that languages' presence in the original model pretraining also contributes much to the quality of the resulting inference. A surprise therefore, is all models' poor performance on English data. We can speculate that Portuguese hits a ``sweet spot'' between language similarity to much of the pretraining data, and size of the task's in-language dataset. But further inspection of the provided dataset's complexity across languages would be required before drawing any conclusion.
\subsection{Supervised Fine-Tuning} 
We were surprised to see how hard it was to train better models than the \xlmr using SFT and Llama or Mistral. These fine-tuning cycles are resource-demanding, and in our prompts, as shown in Appendix \ref{appendix:nlprompts} we provide one prompt per text, requiring the model to return all roles for all entities. To simplify the task, one could request classifications only one entity per prompt. Additionally, applying context optimization might prove beneficial here as well. Testing these options with SFT was beyond our resource allocations.

\section{Conclusion}
We have created a \xlmr-based multilingual model for Entity Framing as a part our contribution to the SemEval 2025 shared task 10 on Multilingual Characterization and Extraction of Narratives from Online News, subtask 1. This model was tested against Supervised Fine-Tuning of Llama and Mistral, and against zero-shot prompting of \cgpt-4o. We found it imperative to extract entity-oriented text segments in order to effectively utilize \xlmr with long documents containing multiple entities each.
\section*{Acknowledgments}
Parts of the work documented in this publication has been carried out within the NorwAI Centre for Research-based Innovation, funded by the Research Council of Norway (RCN), with grant number 309834.\\
 Computations were performed in part on resources provided through Sigma2 -- the national research infrastructure provider for High-Performance Computing and large-scale data storage in Norway.
\bibliography{anthology,custom, jabref_phd}
\appendix
\section*{Appendices}
\clearpage
\onecolumn
\section{Language-Wise Fine-Tuning} \label{appendix:languagewise}

\begin{figure}[H] %
    \centering
    \includegraphics[trim={0pt 200pt 0pt 0pt}, clip, width=0.95\textwidth]{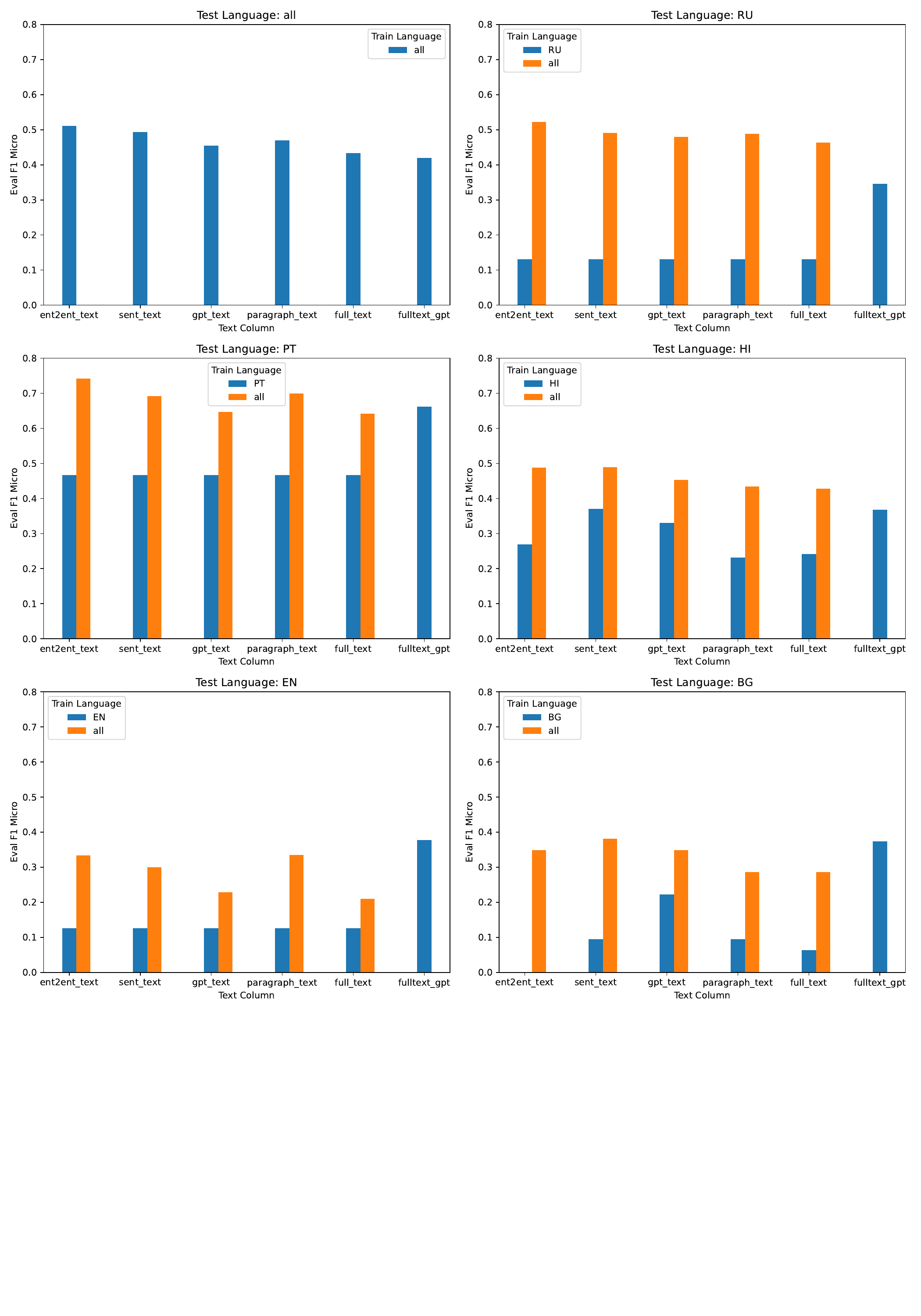} %
    \caption{Benefits from training on all languages. For each language, the results improve substantially when fine-tuning the \xlmr on the entire dataset, as opposed to fine-tuning on the test language only.}
    \label{fig:trainset-contextresults}
\end{figure}

\newpage
\onecolumn
\section{SFT Prompt Template}

\label{appendix:nlprompts}
\lstset{
  basicstyle=\scriptsize\ttfamily,
  numbers=none, %
  showstringspaces=false,
  tabsize=2,
  breaklines=true,
  breakatwhitespace=false,
  frame=single,
  breakautoindent=true,
  columns=fullflexible
}

\begin{lstlisting}[label=prompt_standard:def, basicstyle=\scriptsize\ttfamily]
### Annotation Instructions:
You are given a document that includes various entities along with descriptions of events and actions. Your task is to analyze the 
text and determine the roles each entity plays according to the taxonomy provided below. 

### Taxonomy: 
**Protagonist** 
- Guardian 
- Martyr 
- Peacemaker 
- Rebel 
- Underdog 
- Virtuous 
**Antagonist** 
- Instigator 
- Conspirator 
- Tyrant 
- Foreign Adversary 
- Traitor 
- Spy 
- Saboteur 
- Corrupt 
- Incompetent 
- Terrorist 
- Deceiver 
- Bigot 
**Innocent** 
- Forgotten 
- Exploited 
- Victim 
- Scapegoat 

### Definitions
- **Protagonist**: A central character or force in a positive role.
- **Antagonist**: A character or force in opposition to the protagonist.
- **Innocent**: Entities that are marginalized or victimized without any active role in the conflict.

### New Input:
<<variable input start>>
**LANG: EN**
**Document:**
According to the Gospel of the Global Warming Hoax, 1850-1910 was the coldest period of the past millennium. Yet glaciers were 
retreating rapidly. Now that the planet allegedly has a fever, the retreat has slowed dramatically and even reversed:
Our moonbat rulers canceled the Medieval Warm Period and Little Ice Age for failing to comply with climate ideology. But
preventing glaciers from growing is more difficult than doctoring the historical record to support climate con man <entity>Michael 
Mann</entity>'s spurious hockey stick graph.
Nonetheless, prophet of doom <entity>Al Gore</entity> shouts that "we could lose our capacity for self-governance" if we don't 
surrender still more freedom to Big Government so that it can fix the supposedly broken weather.
On tips from Lyle and Wiggins.
Anyone can join.
Anyone can contribute.
Anyone can become informed about their world.

**Query entities:**
<entity>Michael Mann</entity>
<entity>Al Gore</entity>

### Now for this new document, extract the roles for the following entities:
["Michael Mann", "Al Gore"]
<<variable input end>>
### Output Format
```json
[["entity1", "primary role", ["secondary role 1", "secondary role 2"]],
["entity2", "primary role", ["secondary role 1", ..]]
..]```
"""
\end{lstlisting}

\end{document}